\documentclass[runningheads]{llncs}

\usepackage{eccv}
\usepackage{multirow}
\usepackage{marvosym}

\usepackage{eccvabbrv}

\usepackage{graphicx}
\usepackage{booktabs}

\usepackage[accsupp]{axessibility}  

\usepackage{hyperref}

\usepackage{orcidlink}
\usepackage[ruled,vlined]{algorithm2e}

\begin{document}

\title{TOC-SR: Task-Optimal Compact diffusion for Image Super Resolution} 

\titlerunning{TOC-SR}

\author{Sowmya Vajrala$^{1,*, \textsuperscript{\Letter}}$ \and
Akshay Bankar$^{1,*}$ \and
Manjunath Arveti$^{1,*}$ \and
Shreyas Pandith$^{1}$ \and
Sravanth Kodavanti$^{1}$ \and
Subhajit Sanyal$^{1}$ \and
Amit Unde$^{1}$ \and
Srinivas Miriyala$^{1, \textsuperscript{\Letter}}$ \\
}

\authorrunning{S. Vajrala et al.}

\institute{Samsung Research Institute Bangalore, India}
\maketitle

\setcounter{footnote}{0}
\begingroup{
\let\thefootnote\relax\footnotetext{\textsuperscript{\Letter}{v.lahari@samsung.com, srinivas.soumitri@gmail.com}} 
\let\thefootnote\relax\footnotetext{\textsuperscript{*}Equal Contribution}}

\begin{abstract}

Diffusion models have recently demonstrated strong performance for image restoration tasks, including super-resolution. However, their large model size and iterative sampling procedures make them computationally expensive for practical deployment. In this work, we present \textbf{TOC-SR}, a framework for building efficient one-step super-resolution models by first discovering a compact diffusion backbone. Starting from a sixteen-channel latent diffusion model, we construct parameter-efficient surrogate blocks using feature-wise generative distillation and perform architecture discovery using $\epsilon$-constrained Bayesian Optimization to minimize model complexity while preserving generative fidelity. The resulting compact diffusion backbone achieves a $6.6\times$ reduction in parameters and a $2.8\times$ reduction in GMACs compared to the expanded diffusion model. We then adapt this backbone for super-resolution and distill the diffusion process into a single-step generator. Experiments demonstrate that the proposed approach enables efficient super-resolution while maintaining strong reconstruction quality. 
\end{abstract}

\section{Introduction}
\label{sec:intro}

Single-image super-resolution (SR) aims to reconstruct a high-resolution (HR) image from a low-resolution (LR) observation. The problem is inherently ill-posed, as multiple HR images may correspond to the same LR input. Deep learning has significantly improved SR performance by leveraging large datasets and expressive neural architectures. More recently, diffusion-based generative models have demonstrated strong capability in modeling natural image statistics, enabling high-fidelity restoration that often surpasses traditional regression-based approaches. By progressively refining images through learned denoising dynamics, diffusion models provide a powerful generative prior well suited for ill-posed reconstruction tasks such as SR.

Despite their strong reconstruction ability, diffusion-based SR methods remain computationally expensive. Standard diffusion pipelines require tens of denoising steps and large backbone networks, leading to high latency and memory usage. Recent work has therefore explored distilling diffusion processes into few-step or single-step generators and designing lightweight architectures. While these approaches reduce inference cost, directly training compact one-step SR models often leads to unstable optimization and degraded quality, since the model must simultaneously learn the generative prior and the restoration mapping within limited capacity.

Motivated by this challenge, we explore an alternative design principle: instead of directly constructing compact SR models, we first derive a compact generative backbone that preserves the diffusion prior and then specialize it for restoration. Starting from the latent diffusion architecture of Stable Diffusion 1.5, we revisit the representational capacity of its latent space. The standard model employs a four-channel latent representation, which forms a bottleneck for modeling the high-frequency details required in SR. To alleviate this limitation, we expand the latent representation from four to sixteen channels while maintaining the same spatial compression factor. This increases the expressive capacity of the diffusion prior with only a modest computational overhead ($\sim0.9\times$ increase in parameters and GMACs). As shown in Figure~\ref{fig:fig1}, this expanded model provides a stronger starting point for discovering a compact yet expressive generative backbone.

\begin{figure}
    \centering
    \includegraphics[width=1.0\linewidth]{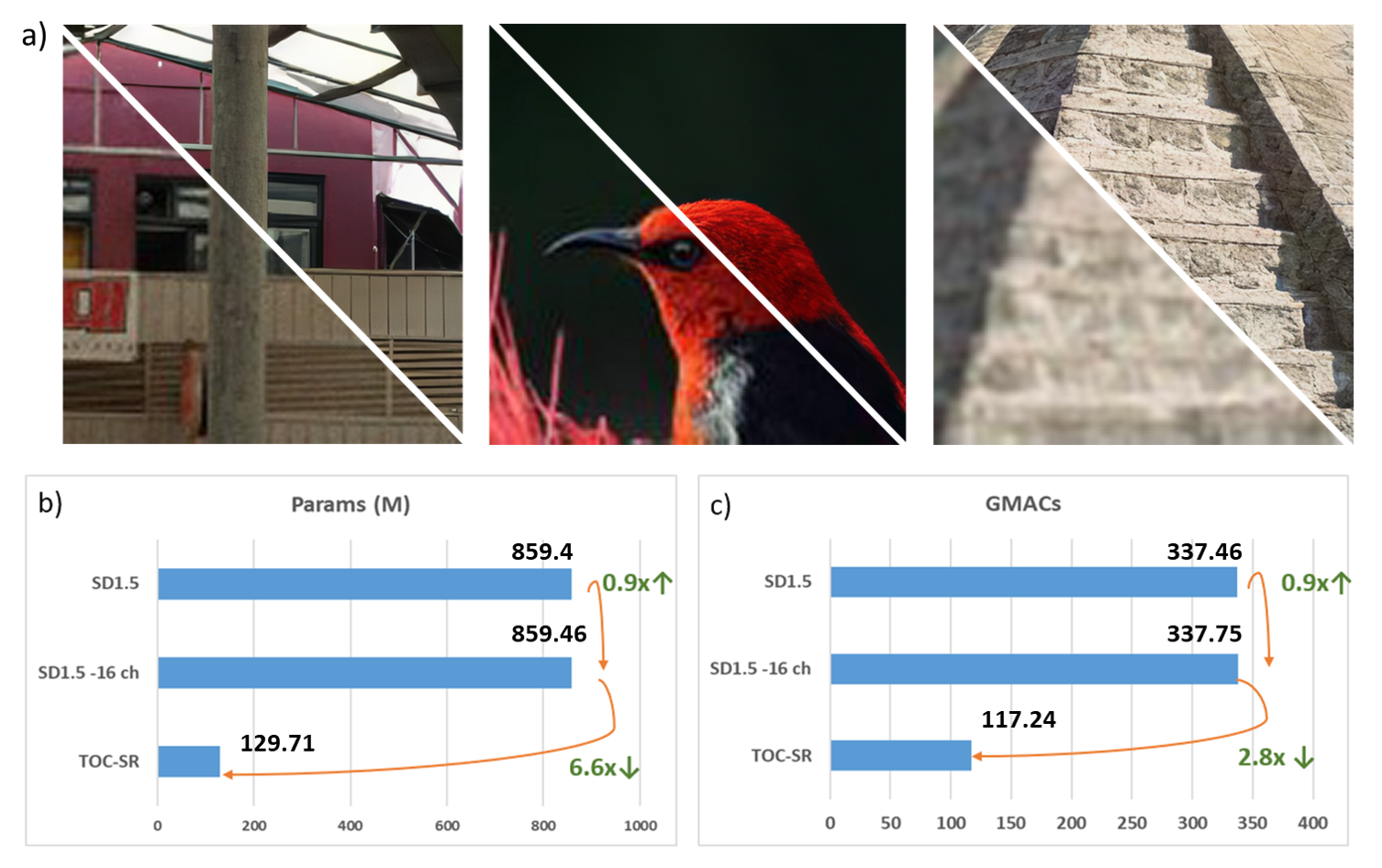}
    \caption{a) Visual Results showing the low quality given as input and high quality images generated by TOC-SR. b) and c) Comparison of Params and GMACs of base SD1.5 model, adapted SD1.5 model and TOC-SR.}
    \label{fig:fig1}
\end{figure}

Building on this representation, we search for a compact diffusion backbone that preserves the teacher model's generative behavior while reducing computational cost. We formulate architecture discovery as a constrained optimization problem and employ $\epsilon$-constrained Bayesian optimization under feature-wise knowledge distillation. Candidate architectures are evaluated based on parameter efficiency while ensuring their distillation loss remains within an acceptable tolerance. This strategy enables systematic discovery of compact architectures that remain faithful to the original diffusion dynamics without requiring task-specific SR training during the search stage.

After identifying the compact backbone, we adapt the model to the SR task through supervised fine-tuning. The resulting diffusion SR model is then distilled into a single-step generator, enabling efficient inference without iterative denoising. As illustrated in Figure~\ref{fig:fig1}, the overall pipeline consists of three stages: latent capacity expansion to alleviate the VAE bottleneck, compact diffusion backbone discovery via $\epsilon$-constrained Bayesian optimization, and task specialization followed by one-step distillation. Starting from the expanded 16-channel diffusion model, the discovered architecture achieves a $6.6\times$ reduction in parameters and a $2.8\times$ reduction in GMACs while retaining strong generative capability.

Overall, we present \textbf{TOC-SR}, an efficient SR framework that combines the generative strength of diffusion models with compact architectures and one-step inference. Our results suggest that discovering compact diffusion priors provides a stable and effective pathway for building high-quality and efficient restoration systems.

\section{Related Works}

\paragraph{Super-Resolution.}
Single-image super-resolution (SR) aims to reconstruct a high-resolution (HR) image from a low-resolution (LR) observation and has been widely studied in computer vision. Early deep learning approaches such as SRCNN~\cite{dong2015image} demonstrated that convolutional neural networks can effectively learn end-to-end LR-to-HR mappings. Subsequent works improved performance by adopting deeper architectures and residual learning, including VDSR~\cite{kim2016accurate},  EDSR~\cite{lim2017enhanced}, and RCAN~\cite{zhang2018image}. While these regression-based approaches achieve strong quantitative performance, they often produce overly smooth textures due to the use of pixel-wise optimization objectives.


\paragraph{Diffusion-based Super-Resolution.}
Diffusion models have recently shown strong performance in image restoration. Methods such as SR3~\cite{saharia2022image} treat super-resolution as a conditional denoising process that progressively refines noisy samples into HR images guided by the LR input. Leveraging strong generative priors, these models can synthesize realistic textures that deterministic networks often fail to recover~\cite{ho2020denoising,rombach2022high}. However, they typically require many denoising steps and large backbones, leading to high computational cost and slow inference.

\paragraph{Efficient Diffusion-based SR.}
To improve the efficiency of diffusion models, recent works explore latent-space modeling, architectural compression, and knowledge distillation. Latent diffusion models~\cite{rombach2022high} reduce computation by performing denoising in compressed feature spaces. More recent methods such as PocketSR~\cite{pocketsr} design lightweight diffusion architectures with generative distillation to retain the capabilities of larger models. However, building compact diffusion backbones that preserve strong generative priors while maintaining reconstruction quality remains challenging.

\paragraph{One-step Super-Resolution.}
Recent work also explores single-step diffusion frameworks to accelerate diffusion-based image restoration and super-resolution. Unlike conventional diffusion models that require many denoising steps, these methods compress the diffusion process into a single forward pass. For example, SinSR~\cite{sinsr}, OSEDiff~\cite{osediff}, and S3Diff~\cite{s3diff} distill the iterative denoising process into efficient one-step models while preserving the generative priors of diffusion models.

\paragraph{EdgeSR and Recent SR Advances.}
Despite recent advances, deploying SR models on resource-constrained devices remains challenging due to the high computational cost of modern architectures. Recent works therefore focus on edge-friendly SR models that balance efficiency and reconstruction quality. For instance, Edge-SD-SR~\cite{noroozi2024edge} proposes a parameter-efficient diffusion framework for low-latency SR, while TinySR~\cite{tinysr}, chen2025adversarial~\cite{chen2025adversarial}, and NanoSD~\cite{sanyal2026nanosd} explore lightweight and compressed diffusion models for efficient on-device deployment.

\section{Formulation}

\subsection{Latent Capacity Expansion from 4 to 16 Channels}

Latent diffusion models such as Stable Diffusion 1.5 operate on a compressed latent representation produced by a variational autoencoder (VAE). Given an input image $x \in \mathbb{R}^{H \times W \times 3}$, the encoder $E(\cdot)$ maps the image to a latent representation $z$ with spatial compression factor $f$:

\begin{equation}
z = E(x), \quad z \in \mathbb{R}^{C \times \frac{H}{f} \times \frac{W}{f}}
\end{equation}

where $C=4$ in the original Stable Diffusion architecture and $f=8$ denotes the spatial downsampling factor. While this representation is sufficient for text-to-image generation, the four-channel latent space introduces a strong information bottleneck for restoration tasks such as super-resolution, where preserving high-frequency structures and fine image details is critical. Recent generative models have therefore begun adopting higher-dimensional latent representations that provide greater capacity for modeling detailed image statistics.

Motivated by this observation, we expand the latent dimensionality from four channels to sixteen channels while preserving the same spatial compression factor. Since no standard Stable Diffusion model with a sixteen-channel latent representation exists, we construct a new latent diffusion backbone by distilling the representational capacity of a high-capacity VAE into a modified Stable Diffusion VAE architecture. Specifically, we use a pretrained sixteen-channel VAE from the FLUX model family as a teacher autoencoder and transfer its latent representation capability into the Stable Diffusion VAE through latent-space distillation.

Let $E_t(\cdot)$ and $D_t(\cdot)$ denote the encoder and decoder of the teacher VAE, producing latent representations

\begin{equation}
z_t = E_t(x), \quad z_t \in \mathbb{R}^{16 \times \frac{H}{f} \times \frac{W}{f}}.
\end{equation}

We modify the Stable Diffusion encoder $E_s(\cdot)$ to produce sixteen latent channels instead of four, while maintaining the same architectural structure. The student encoder therefore produces

\begin{equation}
z_s = E_s(x), \quad z_s \in \mathbb{R}^{16 \times \frac{H}{f} \times \frac{W}{f}}.
\end{equation}

To transfer the latent representation capability from the teacher VAE to the modified student VAE, we employ latent-space distillation. The student encoder is trained to match the teacher latent distribution while simultaneously maintaining accurate image reconstruction through the decoder $D_s(\cdot)$. The resulting objective combines latent distillation with reconstruction supervision:

\begin{equation}
\mathcal{L}_{VAE} =
\lambda_{lat} \| z_s - z_t \|_2^2 +
\lambda_{rec} \| x - D_s(z_s) \|_1 .
\end{equation}

This training procedure enables the modified VAE to produce high-capacity sixteen-channel latent representations while preserving the reconstruction fidelity of the original autoencoder.

After obtaining the sixteen-channel VAE, we adapt the diffusion backbone accordingly. The denoising U-Net in Stable Diffusion is designed to operate on four-channel latent tensors. To support the expanded latent representation, we modify the first and final convolution layers of the U-Net to accept and produce sixteen channels. Let $\epsilon_\theta(\cdot)$ denote the diffusion noise prediction network parameterized by $\theta$. The diffusion training objective remains unchanged and follows the standard denoising score matching formulation:

\begin{equation}
\mathcal{L}_{diff} =
\mathbb{E}_{z, t, \epsilon}
\left[
\|
\epsilon - \epsilon_\theta(z_t, t, c)
\|_2^2
\right],
\end{equation}

where $z_t$ denotes the noisy latent at diffusion timestep $t$, $\epsilon$ is the sampled Gaussian noise, and $c$ represents the conditioning signal.

The modified U-Net and the sixteen-channel VAE are then jointly fine-tuned to ensure that the diffusion model maintains strong generative capability in the expanded latent space. This joint optimization allows the diffusion backbone to adapt to the higher-dimensional latent representation while preserving the generative prior learned from large-scale image data.






\subsection{Construction of a Compact Diffusion Backbone via Surrogate-based $\epsilon$-Constrained Bayesian Optimization}

While the previous subsection establishes a higher-capacity diffusion backbone operating in an expanded latent space, the resulting model remains computationally expensive for practical deployment. Our objective is therefore to derive a compact diffusion model that preserves the generative prior of the teacher while substantially reducing parameter count. Directly training or evaluating thousands of candidate diffusion architectures is computationally prohibitive. To address this challenge, we introduce a surrogate-based architecture exploration framework built upon feature-wise generative knowledge distillation (FwKD) and $\epsilon$-constrained Bayesian Optimization. The proposed framework decomposes the diffusion backbone into stage-wise components, constructs parameter-efficient surrogate variants for each stage, and subsequently searches for compact architectures using a constrained optimization formulation that explicitly preserves generative fidelity.

\begin{figure}
    \centering
    \includegraphics[width=1.0\linewidth]{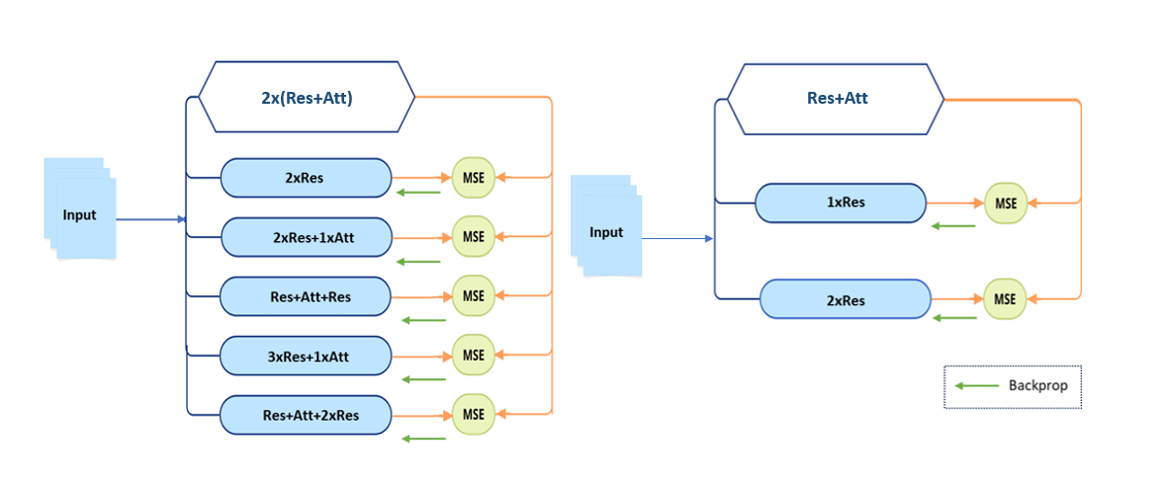}
    \caption{Library of parameter-efficient surrogate blocks for each base block}
    \label{fig:surrogates}
\end{figure}

\subsubsection{Stage-wise Surrogate Construction via Feature-wise Generative Distillation}

Let $\mathcal{T}_{\theta}$ denote the teacher diffusion model obtained in the previous subsection, consisting of a U-Net backbone and a latent autoencoder operating in the sixteen-channel latent space. The U-Net backbone is composed of $S$ sequential stages corresponding to encoder, bottleneck, and decoder blocks. Each stage contains residual and attention modules that collectively define the parameter footprint of the diffusion network.

Rather than directly searching over complete architectures, we construct a library of parameter-efficient surrogate blocks (see Figure ~\ref{fig:surrogates}) for each stage through structured architectural reductions. Specifically, we derive a set of candidate variants by systematically removing or simplifying attention modules, residual blocks, or channel expansions while strictly preserving input-output tensor dimensionality. This constraint ensures that all candidate blocks remain interchangeable within the diffusion backbone without requiring additional adapters or tensor reshaping.

Let $B_i$ denote the teacher block at stage $i$. For each stage we construct a set of candidate surrogate blocks

\begin{equation}
\mathcal{B}_i = \{B_{i,1}, B_{i,2}, \dots, B_{i,K_i}\},
\label{eq:surrogate_blocks}
\end{equation}

where each $B_{i,j}$ represents a parameter-reduced architectural variant derived from $B_i$. These candidates are obtained through structured pruning and module reduction strategies that target parameter-heavy components while preserving functional compatibility.

Training each candidate architecture end-to-end would require repeated diffusion training, which is infeasible. Instead, we calibrate each surrogate block independently using feature-wise generative distillation. Let $F_i$ denote the input feature tensor to stage $i$ extracted from the teacher network. The outputs of the teacher and student blocks are given by

\begin{equation}
O_i^{T} = B_i(F_i), \quad O_{i,j}^{S} = B_{i,j}(F_i).
\label{eq:block_outputs}
\end{equation}

Each surrogate block is calibrated by minimizing a feature reconstruction objective:

\begin{equation}
\mathcal{L}_{\text{FwKD}}^{(i,j)} =
\mathbb{E}_{F_i}
\left[
\| O_{i,j}^{S} - O_i^{T} \|_2^2
\right].
\label{eq:fwkd}
\end{equation}

Importantly, this procedure does not involve full diffusion training. Instead, we utilize a calibration subset comprising $25\%$ of the available data to expose surrogate blocks to representative latent features produced by the teacher model. Since each surrogate block is optimized independently, the calibration process is computationally lightweight and can be parallelized across stages and candidate variants. After calibration, every surrogate block approximates the functional behavior of its teacher counterpart while possessing a reduced parameter footprint.

\subsubsection{Plug-and-Play Assembly of Candidate Diffusion Architectures}

Once surrogate blocks are calibrated, they can be assembled into complete diffusion backbones without additional training. A candidate architecture is defined by a discrete decision vector

\begin{equation}
\mathbf{a} = [a_1, a_2, \dots, a_S],
\label{eq:architecture_vector}
\end{equation}

where $a_i \in \{1, \dots, K_i\}$ selects a surrogate block from $\mathcal{B}_i$ for stage $i$.

Because all surrogate variants preserve tensor shapes, substituting blocks according to $\mathbf{a}$ yields a structurally valid diffusion backbone. The resulting model, denoted $\mathcal{M}(\mathbf{a})$, can therefore be instantiated simply by replacing teacher blocks with the corresponding calibrated surrogates. This plug-and-play property enables rapid evaluation of candidate architectures without retraining.

The parameter complexity of the resulting architecture is given by

\begin{equation}
f_{\text{param}}(\mathbf{a}) = \text{Params}(\mathcal{M}(\mathbf{a})).
\label{eq:param}
\end{equation}

To quantify the deviation from the teacher's generative behavior, we measure the discrepancy between noise predictions of the candidate and teacher models across the calibration dataset. Let $\epsilon_{\theta}(z_t, t, c)$ denote the teacher noise prediction and $\epsilon_{\phi(\mathbf{a})}(z_t, t, c)$ the prediction of candidate architecture $\mathcal{M}(\mathbf{a})$. We define the generative fidelity objective as

\begin{equation}
f_{\text{acc}}(\mathbf{a}) =
\mathbb{E}_{z_t,t,c}
\left[
\|
\epsilon_{\phi(\mathbf{a})}(z_t,t,c) -
\epsilon_{\theta}(z_t,t,c)
\|_2^2
\right].
\label{eq:accuracy}
\end{equation}

This metric measures how closely the candidate architecture reproduces the denoising behavior of the teacher diffusion model.

\subsubsection{$\epsilon$-Constrained Bayesian Optimization}

Selecting an optimal architecture naturally yields a multi-objective problem involving model complexity and generative fidelity:

\begin{equation}
\min_{\mathbf{a}} \; (f_{\text{param}}(\mathbf{a}), f_{\text{acc}}(\mathbf{a})).
\label{eq:moop}
\end{equation}

Standard approaches for multi-objective optimization include weighted-sum formulations and Lagrangian relaxation. However, these approaches introduce instability due to objective scaling and require careful tuning of trade-off parameters. Furthermore, our goal is not to obtain a full Pareto frontier but rather to identify a compact model that preserves the generative prior within an acceptable tolerance.

We therefore reformulate the problem as an $\epsilon$-constrained single-objective optimization:

\begin{equation}
\min_{\mathbf{a}} f_{\text{param}}(\mathbf{a})
\label{eq:eps_obj}
\end{equation}

subject to

\begin{equation}
f_{\text{acc}}(\mathbf{a}) \leq \epsilon.
\label{eq:eps_constraint}
\end{equation}

The parameter $\epsilon$ controls the maximum allowable deviation from the teacher model. Smaller values of $\epsilon$ enforce stricter fidelity constraints, producing architectures that closely mimic the teacher but offer limited compression. Larger values allow more aggressive parameter reduction at the cost of increased approximation error.

Evaluating $f_{\text{acc}}(\mathbf{a})$ requires running diffusion inference over the calibration set and is therefore computationally expensive. To efficiently explore the architecture space, we employ Bayesian Optimization (BO). A Gaussian Process surrogate model approximates the objective and constraint functions based on previously evaluated architectures. Candidate architectures are selected using an acquisition function that balances exploration and exploitation while satisfying the $\epsilon$ constraint.

To systematically explore the trade-off between model compactness and fidelity, we perform BO under multiple $\epsilon$ values sampled from a predefined grid

\begin{equation}
\epsilon \in \{\epsilon_1, \epsilon_2, \dots, \epsilon_K\}.
\label{eq:epsilon_grid}
\end{equation}

Each optimization run produces a candidate architecture satisfying the fidelity constraint for that particular tolerance level. This procedure yields a sequence of models spanning different compression regimes. From this set we select the architecture whose parameter count is closest to our target model size of approximately $130$M parameters.

\subsubsection{Final End-to-End Fine-tuning}

Although surrogate calibration ensures approximate functional alignment between candidate architectures and the teacher, assembling surrogate blocks can introduce minor inconsistencies across stages. To correct these accumulated discrepancies, the selected compact architecture is finally fine-tuned end-to-end using the standard diffusion denoising objective:

\begin{equation}
\mathcal{L}_{\text{diff}} =
\mathbb{E}_{z_t,t,c}
\left[
\|
\epsilon - \epsilon_{\phi}(z_t,t,c)
\|_2^2
\right].
\label{eq:diff_loss}
\end{equation}

This final training stage restores full generative capability while preserving the structural efficiency discovered by the architecture search.

\begin{algorithm}[t]
\caption{Surrogate-based $\epsilon$-Constrained Diffusion Architecture Search}
\textbf{Input:} Teacher diffusion model $\mathcal{T}_{\theta}$, calibration dataset $\mathcal{D}_{cal}$, $\epsilon$ grid $\{\epsilon_k\}$  
\textbf{Output:} Compact diffusion backbone $\mathcal{M}^*$  

1: Decompose teacher U-Net into $S$ stages $\{B_i\}_{i=1}^{S}$  

2: For each stage $i$, generate surrogate candidates $\mathcal{B}_i = \{B_{i,j}\}$ via structured reductions  

3: \textbf{for each} candidate block $B_{i,j}$ \textbf{do}  

4: \quad Calibrate using FwKD on $\mathcal{D}_{cal}$ minimizing $\mathcal{L}_{\text{FwKD}}^{(i,j)}$  

5: \textbf{end for}

6: Define architecture vector $\mathbf{a}=[a_1,\dots,a_S]$

7: \textbf{for each} $\epsilon_k$ in grid \textbf{do}

8: \quad Initialize Bayesian Optimization model  

9: \quad \textbf{repeat}

10: \qquad Propose architecture $\mathbf{a}$  

11: \qquad Assemble model $\mathcal{M}(\mathbf{a})$ using surrogate blocks  

12: \qquad Evaluate $f_{\text{param}}(\mathbf{a})$ and $f_{\text{acc}}(\mathbf{a})$

13: \qquad Update BO surrogate  

14: \quad \textbf{until} convergence

15: \quad Record best feasible architecture for $\epsilon_k$

16: \textbf{end for}

17: Select architecture $\mathcal{M}^*$ closest to target parameter budget  

18: Fine-tune $\mathcal{M}^*$ using diffusion objective $\mathcal{L}_{diff}$  

19: \textbf{return} $\mathcal{M}^*$
\end{algorithm}

\subsection{Adapting the Compact Diffusion Backbone for One-Step Super-Resolution}

After identifying the compact sixteen-channel diffusion backbone in the previous subsection, the next step is to specialize the model for the super-resolution task and subsequently convert the iterative diffusion process into a single-step generator. Our objective is to preserve the generative prior learned by the diffusion backbone while enabling efficient inference suitable for deployment. To achieve this, we follow a two-stage training procedure: first, the compact diffusion backbone is adapted for image-to-image super-resolution through conditional diffusion training. Second, the resulting super-resolution diffusion model is distilled into a single-step generator that directly predicts the high-resolution output.

\subsubsection{Conditional Diffusion Training for Super-Resolution}

We adopt a latent diffusion formulation in which both the low-resolution (LR) input and the target high-resolution (HR) image are represented in the latent space of the VAE. Let $x_h$ denote the ground-truth HR image and $x_l$ the corresponding LR image obtained through bicubic downsampling by a factor of $3\times$. The HR image is encoded into the latent space using the VAE encoder introduced earlier:

\begin{equation}
z_h = E(x_h),
\label{eq:hr_latent}
\end{equation}

where $z_h \in \mathbb{R}^{16 \times \frac{H}{8} \times \frac{W}{8}}$ denotes the sixteen-channel latent representation.

Following the standard diffusion formulation, Gaussian noise is progressively added to the HR latent representation over $T$ diffusion steps. Let $z_t$ denote the noisy latent at timestep $t$ obtained through the forward diffusion process

\begin{equation}
z_t = \sqrt{\alpha_t} z_h + \sqrt{1-\alpha_t}\,\epsilon,
\quad \epsilon \sim \mathcal{N}(0, I).
\label{eq:forward_diffusion_sr}
\end{equation}

The U-Net denoising network is trained to predict the injected noise $\epsilon$ conditioned on the LR image. To obtain the LR conditioning signal, the LR image is first upsampled to the HR resolution and then encoded through the same VAE encoder

\begin{equation}
z_l = E(\text{Upsample}(x_l)).
\label{eq:lr_latent}
\end{equation}

The denoising network therefore learns the conditional mapping

\begin{equation}
\epsilon_{\phi}(z_t, t, z_l),
\label{eq:conditional_denoiser}
\end{equation}

where $\phi$ denotes the parameters of the compact diffusion backbone obtained from the previous section.

The training objective follows the standard denoising score-matching loss

\begin{equation}
\mathcal{L}_{SR} =
\mathbb{E}_{z_h,t,\epsilon}
\left[
\|
\epsilon -
\epsilon_{\phi}(z_t, t, z_l)
\|_2^2
\right].
\label{eq:sr_diffusion_loss}
\end{equation}

This training procedure enables the diffusion model to progressively refine the noisy latent representation toward the HR latent conditioned on the LR observation. During inference, the model starts from Gaussian noise and iteratively performs denoising steps guided by the LR conditioning latent until a clean HR latent is obtained, which is subsequently decoded through the VAE decoder to produce the super-resolved image.

\subsubsection{One-Step Distillation for Efficient Super-Resolution}

Although diffusion-based SR models produce high-quality results, iterative denoising across many timesteps results in high computational cost. To enable efficient inference, we distill the multi-step diffusion SR model into a single-step generator following the step distillation paradigm used in recent diffusion acceleration methods.

Let $\epsilon_{\phi}$ denote the trained multi-step diffusion model obtained in the previous stage. The goal of step distillation is to train a student network $\epsilon_{\psi}$ that approximates the overall denoising trajectory of the teacher diffusion process in a single step. Specifically, given the LR conditioning latent $z_l$ and an initial noisy latent $z_T$, the student network directly predicts the clean latent representation

\begin{equation}
\hat{z}_0 = G_{\psi}(z_T, z_l),
\label{eq:onestep_generator}
\end{equation}

where $G_{\psi}$ denotes the one-step generator derived from the diffusion backbone.

To supervise the student model, we use the HR latent produced by the teacher diffusion model after performing the full denoising process. Let $z_0^{T}$ denote the final latent obtained from the teacher diffusion model. The student network is trained to regress this target in a single step using the loss

\begin{equation}
\mathcal{L}_{distill} =
\mathbb{E}_{z_T,z_l}
\left[
\|
G_{\psi}(z_T,z_l) - z_0^{T}
\|_2^2
\right].
\label{eq:distillation_loss}
\end{equation}

After training, the student model is capable of directly predicting the clean HR latent from a single noisy latent conditioned on the LR input. The predicted latent is then passed through the VAE decoder to obtain the final super-resolved image. 

By combining the compact diffusion backbone obtained in the previous subsection with the one-step distillation procedure described above, we obtain the final \textbf{TOC-SR} model. The resulting architecture preserves the generative prior learned from diffusion training while enabling efficient single-step super-resolution inference.

\begin{table*}[t]
\centering
\caption{Quantitative comparison on the DIV2K validation set~\cite{agustsson2017ntire}. Best, second, and third results are highlighted in red, blue, and green.}
\label{table:sr1}
\begin{tabular}{lccccccccc}
\toprule
Method & PSNR$\uparrow$ & LPIPS$\downarrow$ & FID$\downarrow$ & NIQE$\downarrow$ & MUSIQ$\uparrow$ & Steps & MACs(G) & Params(M) \\
\midrule
StableSR~\cite{stablesr} & 23.26 & 0.311 & 24.44 & 4.75 & 65.92 & 200 & 79940 & 1410 \\
DiffBIR~\cite{diffbir} & 23.64 & 0.352 & 30.72 & 4.70 & 65.81 & 50 & 24234 & 1717 \\
SeeSR~\cite{seesr} & 23.68 & 0.319 & 25.90 & 4.81 & 68.67 & 50 & 65857 & 2524 \\
ResShift~\cite{resshift} & \textcolor{red}{24.65} & 0.334 & 36.11 & 6.82 & 61.09 & 15 & 5491 & 119 \\
SinSR~\cite{sinsr} & \textcolor{blue}{24.41} & 0.324 & 35.57 & 6.02 & 62.82 & 1 & 2649 & 119 \\
OSEDiff~\cite{osediff} & 23.72 & 0.294 & 26.32 & 4.71 & 67.97 & 1 & 2265 & 1775 \\
S3Diff~\cite{s3diff} & 23.52 & \textcolor{green}{0.258} & \textcolor{red}{19.66} & 4.74 & 68.01 & 1 & 2621 & 1327 \\
TinySR~\cite{tinysr} & -- & 0.279 & \textcolor{green}{22.94} & \textcolor{red}{4.15} & \textcolor{blue}{69.90} & 1 & \textcolor{green}{427} & 341 \\
Edge-SD-SR~\cite{edgesdsr} & 24.10 & \textcolor{blue}{0.249} & 25.37 & -- & \textcolor{green}{69.58} & 1 & -- & 169 \\
PocketSR~\cite{pocketsr} & 23.85 & 0.280 & 25.25 & \textcolor{green}{4.41} & 66.38 & 1 & \textcolor{red}{225} & 146 \\
Chen et al.~\cite{chen2025adversarial} & 23.74 & 0.285 & 25.52 & \textcolor{blue}{4.36} & 68.00 & 1 & 496 & 456 \\
\midrule
\textbf{TOC-SR (Ours)} & \textcolor{green}{24.26} & \textcolor{red}{0.242} & \textcolor{blue}{22.36} & 4.56 & \textcolor{red}{70.14} & 1 & \textcolor{blue}{273} & 130 \\
\bottomrule
\end{tabular}
\end{table*}

\section{Experiments}
\begin{figure*}[ht]
    \centering
    \includegraphics[width=1\linewidth]{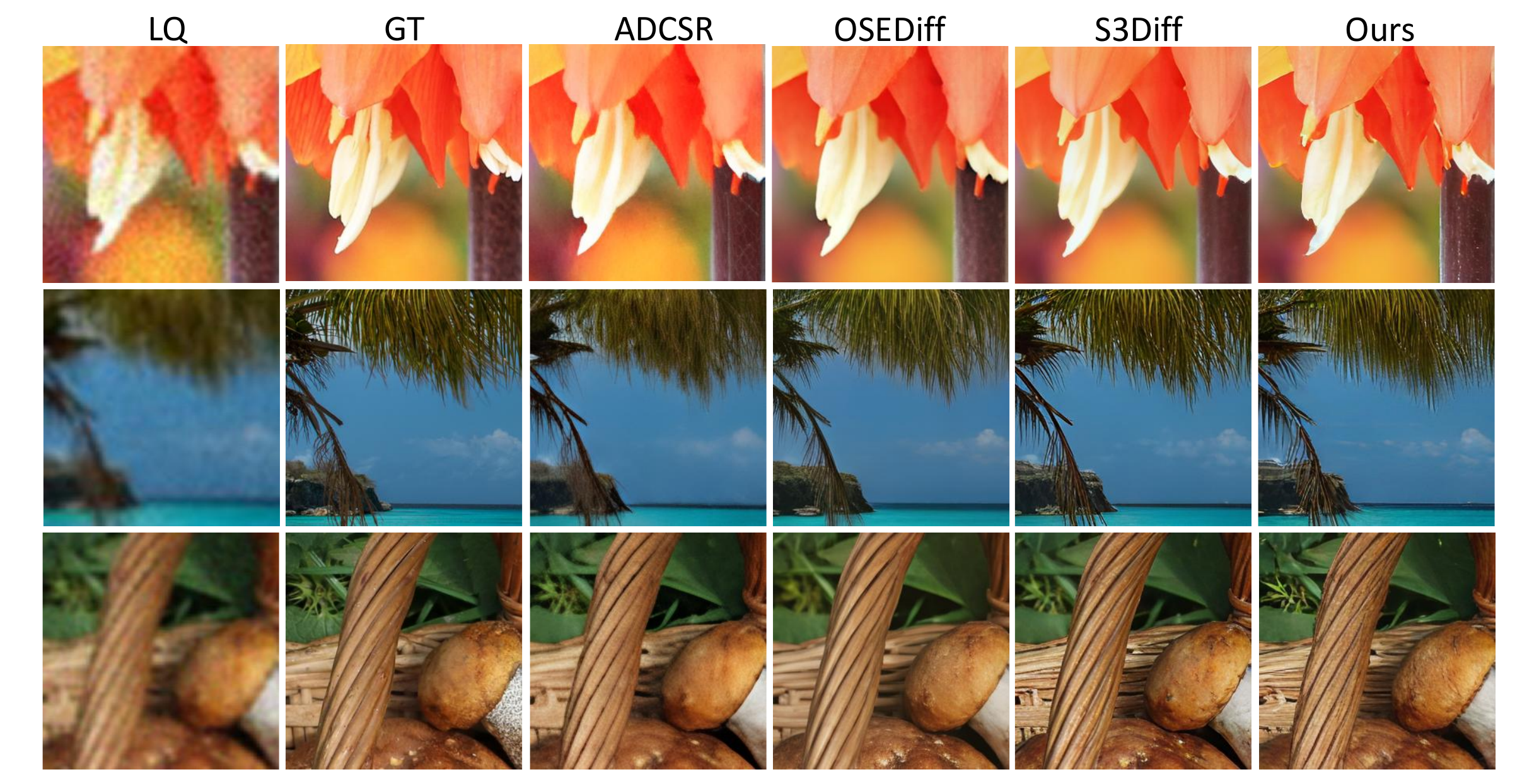}
    \caption{\textbf{Qualitative comparison on $\times4$ super-resolution.}
We compare TOC-SR against representative diffusion SR baselines.
TOC-SR better preserves fine textures and thin structures while avoiding over-smoothing and unnatural micro-textures. Please zoom in for details.}
    \label{fig:sr_results}
\end{figure*}

\begin{table*}[t]
\centering
\caption{Quantitative comparison on real-world datasets. Best, second, and third results are highlighted in red, blue, and green.}
\label{table:sr2}
\small
\begin{tabular}{lcccccccc}
\toprule
& \multicolumn{4}{c}{DRealSR~\cite{drealsr}} & \multicolumn{4}{c}{RealSR~\cite{realsr}} \\
\cmidrule(lr){2-5} \cmidrule(lr){6-9}
Method & PSNR$\uparrow$ & LPIPS$\downarrow$ & FID$\downarrow$ & MUSIQ$\uparrow$ 
       & PSNR$\uparrow$ & LPIPS$\downarrow$ & FID$\downarrow$ & MUSIQ$\uparrow$ \\
\midrule
SinSR~\cite{sinsr} & \textcolor{blue}{28.36} & 0.366 & 170.5 & 55.33 
                   & \textcolor{red}{26.28} & 0.318 & 135.9 & 60.80 \\
OSEDiff~\cite{osediff} & 27.92 & 0.296 & 135.3 & 64.65 
                       & 25.15 & 0.292 & 123.4 & 69.09 \\
S3Diff~\cite{s3diff} & 27.39 & 0.312 & \textcolor{red}{119.2} & 64.16 
                     & 25.19 & \textcolor{red}{0.270} & \textcolor{red}{110.3} & 67.92 \\
Edge-SD-SR~\cite{edgesdsr} & -- & \textcolor{red}{0.292} & -- & 55.66 
                           & -- & 0.278 & -- & 65.20 \\
Chen et al.~\cite{chen2025adversarial} & \textcolor{green}{28.10} & 0.304 & \textcolor{green}{134.0} & \textcolor{blue}{66.26} 
                                        & 25.47 & 0.288 & 118.4 & \textcolor{blue}{69.90} \\
TinySR~\cite{tinysr} & 27.48 & 0.311 & 146.7 & \textcolor{green}{65.36} 
                     & 24.79 & \textcolor{green}{0.280} & \textcolor{green}{118.0} & \textcolor{green}{69.78} \\
PocketSR~\cite{pocketsr} & 28.05 & \textcolor{blue}{0.296} & -- & 63.85 
                         & \textcolor{green}{25.47} & \textcolor{blue}{0.271} & -- & 67.07 \\
\midrule
\textbf{TOC-SR (Ours)} & \textcolor{red}{28.91} & \textcolor{green}{0.301} & \textcolor{blue}{129.7} & \textcolor{red}{66.92} 
                       & \textcolor{blue}{25.73} & 0.286 & \textcolor{blue}{115.4} & \textcolor{red}{70.31} \\
\bottomrule
\end{tabular}
\end{table*}
\subsection{Compact Diffusion Backbone Discovery}

We now evaluate the effectiveness of the proposed surrogate-based $\epsilon$-constrained search in discovering compact diffusion backbones. Starting from the sixteen-channel diffusion model described in Section~3.1, we construct parameter-efficient surrogate variants for each stage of the U-Net using the feature-wise calibration procedure introduced in Section~3.2. In practice, we design $4$ surrogate alternatives for each encoder and decoder stage and $6$ alternatives for the bottleneck stage through structured reductions of residual and attention components while preserving tensor dimensionality. This results in a total architecture search space of approximately $3.2 \times 10^{4}$ feasible diffusion backbones.

Architecture search is performed using $\epsilon$-constrained Bayesian Optimization as defined in Eqs.~\ref{eq:eps_obj}--\ref{eq:eps_constraint}. We explore a grid of $8$ $\epsilon$ values to systematically control the fidelity--compression trade-off. For each $\epsilon$, the Bayesian optimizer runs for $40$ iterations, resulting in approximately $320$ evaluated candidate architectures in total. Additional implementation details and intermediate search statistics are provided in the supplementary material.

Among the feasible architectures discovered during the search, we select the model whose parameter count best satisfies our deployment budget. The selected architecture is subsequently fine-tuned end-to-end using the diffusion objective in Eq.~\ref{eq:diff_loss} to restore full generative capability. Compared to the expanded sixteen-channel diffusion backbone introduced in Section~3.1, the resulting compact model achieves a \textbf{$6.6\times$ reduction in parameters} and a \textbf{$2.8\times$ reduction in GMACs}. We refer to this compact backbone as \textbf{TOC-SR}, which serves as the foundation for the super-resolution experiments presented next.

\subsection{Super Resolution Performance}

\subsubsection{Datasets and Metrics}
\label{subsec:datasets}
Following prior work, we use a mixture of large-scale natural images and face images for training. Specifically, we train on LSDIR~\cite{li2023lsdir}, the first 10K images from FFHQ~\cite{karras2019stylegan}, and DIV2K-Train~\cite{agustsson2017ntire}. To construct paired low-quality (LQ) and high-quality (HQ) training data, we adopt the synthetic degradation pipeline of Real-ESRGAN~\cite{wang2021realesrgan}.
For evaluation on real-world benchmarks, we report results on DIV2K-Val~\cite{agustsson2017ntire}, RealSR~\cite{realsr}, and DRealSR~\cite{drealsr}. \cite{li2023lsdir,karras2019stylegan,agustsson2017ntire,wang2021realesrgan,realsr,drealsr}

To provide a comprehensive assessment, we employ both full-reference and no-reference metrics. We report PSNR as a reference-based fidelity measure, and LPIPS~\cite{zhang2018lpips} as a perceptual similarity metric. Additionally, we use NIQE~\cite{zhang2015niqe} and MUSIQ~\cite{ke2021musiq} as no-reference image quality measures. \cite{zhang2018lpips,zhang2015niqe,ke2021musiq}






\subsubsection{Results}
\label{subsec:results}

\paragraph{Quantitative comparison on DIV2K-Val.}
Table ~\ref{table:sr1}  compares TOC-SR against recent diffusion-based and compact SR baselines on DIV2K-Val.
TOC-SR achieves 24.26 dB PSNR and the best LPIPS 0.242 among the listed methods, indicating superior perceptual fidelity at the same $\times4$ scaling.
In addition, TOC-SR attains a strong FID of 22.36 while using a single-step inference.
Importantly, TOC-SR improves the quality--efficiency trade-off: compared to multi-step diffusion SR approaches (e.g., StableSR, DiffBIR, SeeSR), it reduces the number of denoising evaluations from tens to hundreds down to 1 step, while remaining competitive (and often stronger) in perceptual metrics.

\paragraph{Real-world benchmarks (RealSR and DRealSR).}
As shown in Table~\ref{table:sr2}, TOC-SR generalizes well to real degradations.
On DRealSR, TOC-SR achieves the highest PSNR (28.91) and best MUSIQ (66.92), indicating strong restoration fidelity and improved no-reference perceptual quality.
On RealSR, TOC-SR reaches 25.73 PSNR and 70.31 MUSIQ, outperforming other compact diffusion baselines in overall perceptual quality.
While some methods achieve lower LPIPS on specific datasets (e.g., S3Diff on RealSR), TOC-SR provides a stronger overall balance across reference-based fidelity (PSNR) and perceptual realism (FID/MUSIQ).

\paragraph{Qualitative comparison.}
Figure~\ref{fig:sr_results} shows representative visual comparisons.
Across challenging structures (fine textures, thin edges, and repeated patterns), TOC-SR reconstructs sharper and more coherent details while avoiding common diffusion artifacts such as over-smoothing and texture hallucination.
In particular, compared to compact baselines (e.g., chen2025adversarial, OSEDiff, S3Diff), TOC-SR better preserves local contrast and restores structured regions with fewer ringing/blur artifacts, aligning with its gains in LPIPS and MUSIQ.

\subsection{Ablation Study}
\label{subsec:ablation}

We ablate the vae channel capacity under the same training pipeline. Table~\ref{table:sr3} reports quantitative comparisons on DIV2K-Val. For the proposed U-Net, switching from 16ch to 4ch reduces PSNR from 24.26 to 23.24 and worsens LPIPS from 0.242 to 0.291.These results indicate that latent capacity is a primary bottleneck for SR quality: a higher-dimensional latent better preserves fine structures and enables the denoiser to reconstruct high-frequency details more faithfully.






\begin{table}[t]
\centering
\caption{\textbf{Ablation on VAE latent channels (DIV-2K Val \cite{agustsson2017ntire}, $\times4$).}}
\label{table:sr3}

\setlength{\tabcolsep}{4pt}
\renewcommand{\arraystretch}{0.9}

\footnotesize
\begin{tabular}{lcccccc}
\toprule
Method & PSNR$\uparrow$ & LPIPS$\downarrow$ & FID$\downarrow$ & NIQE$\downarrow$ & MUSIQ$\uparrow$ & Steps \\
\midrule

TOC-SR (ours) & \textcolor{blue}{24.26} & \textcolor{red}{0.242} & \textcolor{red}{22.36} & \textcolor{blue}{4.56} & \textcolor{blue}{70.14} & 1 \\

4-Channel model & 23.24 & \textcolor{green}{0.291} & 27.37 & 5.10 & \textcolor{green}{66.53} & 1 \\

\bottomrule
\end{tabular}
\end{table}

\section{Conclusion}

In this work, we presented \textbf{TOC-SR}, a framework for constructing efficient super-resolution models from compact diffusion backbones. We first expanded the latent representation of Stable Diffusion to sixteen channels and introduced a surrogate-based architecture exploration strategy to discover parameter-efficient diffusion models using $\epsilon$-constrained Bayesian Optimization. The resulting backbone achieves significant reductions in model complexity while preserving the generative prior of the teacher model. By adapting this compact diffusion backbone for super-resolution and performing one-step distillation, we obtain an efficient super-resolution model capable of high-quality reconstruction with reduced computational cost. 

\newpage

%
%
\bibliographystyle{splncs04}
\bibliography{main}
\newpage

\section{Supplementary Material}

\appendix
\renewcommand{\thefigure}{S\arabic{figure}}
\renewcommand{\thetable}{S\arabic{table}}
\renewcommand{\thealgocf}{S\arabic{algocf}}
\section{Latent Capacity Expansion and VAE Distillation}

\begin{figure}
    \centering
    \includegraphics[width=\linewidth]{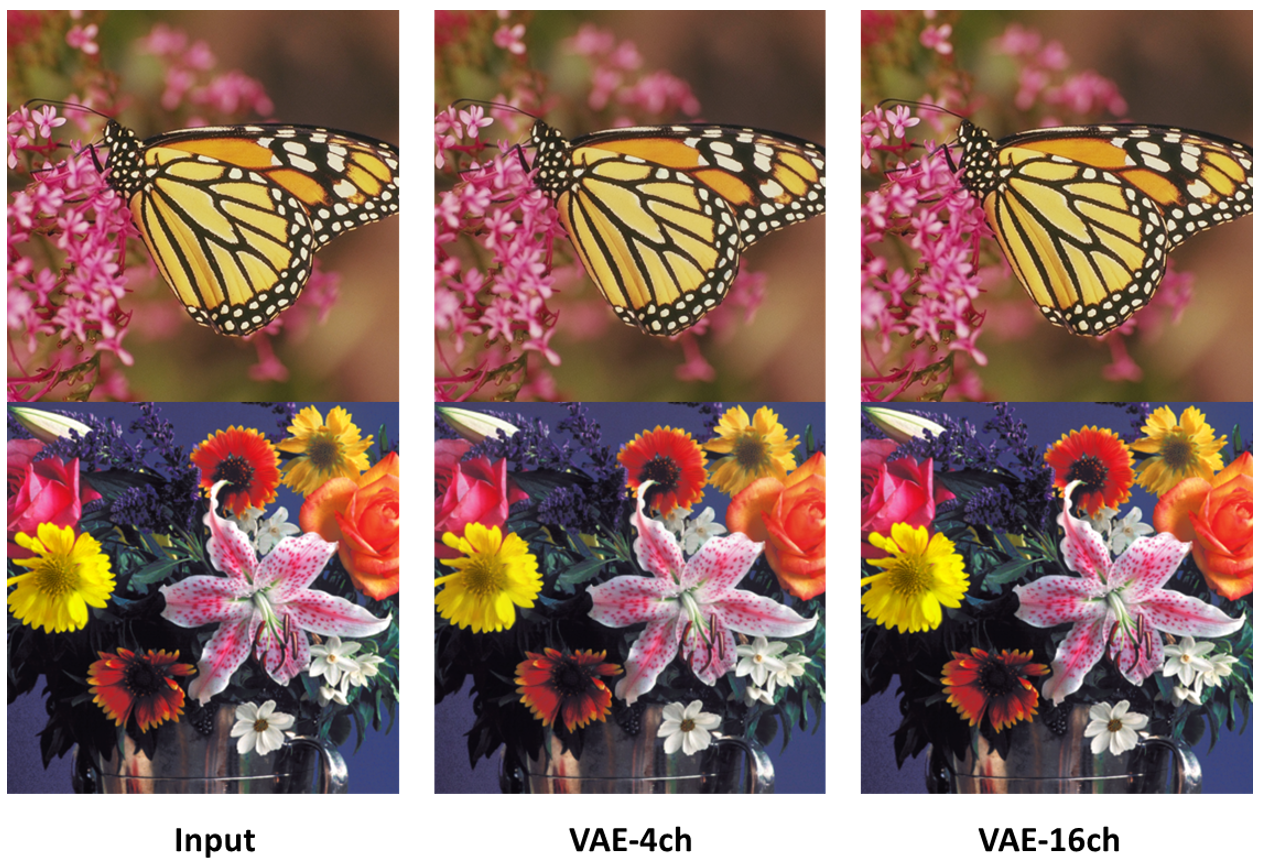}
    \caption{Comparison of Image Reconstruction capabilities of 4 and 16-channel VAEs}
    \label{fig:vae_recon}
\end{figure}

The original Stable Diffusion architecture operates on a four-channel latent representation, which introduces a strong information bottleneck for restoration tasks such as super-resolution. While this representation is sufficient for generative image synthesis, it limits the model’s ability to capture fine textures and high-frequency structures required for accurate reconstruction. To alleviate this limitation, we expand the latent dimensionality from four to sixteen channels while maintaining the same spatial compression factor.

\begin{table}[]
\caption{Ablation on adapting the 4-channel SD 1.5 base model to 16-channel backbone. \textbf{SD1.5\_4ch\_base} is the SD 1.5 model with 859.395M parameters, and \textbf{SD1.5\_16ch\_v2} is the final selected backbone with 859.46 parameters, resulting in 0.008\% excess parameters.}
\centering
\label{tab:conv_adapter}
\footnotesize
\begin{tabular}{llll}
\toprule
\textbf{Model}                            & \textbf{Description}                      & \textbf{Params(M)} & \textbf{Loss} \\
\midrule
\multirow{2}{*}{\textbf{SD1.5\_4ch\_base}}           & Conv\_in = (Conv1(4--\textgreater{}320))   & 859.395                    &               \\
                                          & Conv\_out = (Conv1(320--\textgreater{}4)) \\
\multirow{2}{*}{\textbf{SD1.5\_16ch\_v1}} & Conv\_in = (Conv1(16--\textgreater{}4),     & 859.43                     & 0.7598        \\
                                          & Conv2(4--\textgreater{}320))  \\
                                          & Conv\_out = (Conv1(320--\textgreater{}16))\\
\multirow{2}{*}{\textbf{SD1.5\_16ch\_v2}} & Conv\_in = (Conv1(16--\textgreater{}320))  & 859.46                     & 0.084         \\
                                          & Conv\_out = (Conv1(320--\textgreater{}16))   \\
\multirow{2}{*}{\textbf{SD1.5\_16ch\_v3}} & Conv\_in = (Conv1(16--\textgreater{}8),    & 859.398                     & 0.7642        \\
                                          & Conv2(8--\textgreater{}4),              \\
                                          & Conv3(4--\textgreater{}320))             \\
                                          & Conv\_out = (Conv1(320--\textgreater{}4),   \\
                                          & Conv1(4--\textgreater{}8),                \\
                                          & Conv3(8--\textgreater{}16))             \\
                                          
\bottomrule
\end{tabular}
\end{table}


\begin{table}[]
\caption{Quantitative Comparison between 4 and 16-channel VAEs}
\centering
\label{tab:vae_comp}
\begin{tabular}{llll}
\toprule
                  & \textbf{FID} & \textbf{PSNR} & \textbf{SSIM} \\
                  \midrule
\textbf{4ch VAE}  & 7.939       & 28.8277       & 0.7946        \\
\textbf{16ch VAE} & 7.114        & 29.0284       & 0.9188       \\
\bottomrule
\end{tabular}
\end{table}

Increasing the channel capacity improves the representational power of the latent space with only a modest computational overhead, enabling the diffusion backbone to better model detailed image statistics. Notably, recent generative models such as FLUX also adopt higher-dimensional latent representations to increase generative capacity. Once the VAE distillation is done, the U-Net is also modified to accept and generate 16 channels. This is done by changing the input and output of the first and last Convolution operations of the U-Net. 

To adapt the 4-channel U-Net model to 16-channel, several options were explored, ranging from increasing the input and output channels of start and end convolutions to replacing start and end Convolutions with multiple Convolutions that gradually increase the number of channels, as shown in Table ~\ref{tab:conv_adapter}. All these models were fine-tuned end-to-end using the diffusion loss as an objective (Eq. 16 in the main paper). Among these, the model where input and output channels of start and end convolutions are increased to 16 was found to have the least loss and is therefore selected as the final 16-channel U-Net backbone model. The resultant U-Net is then fine-tuned along with the 16-channel VAE as described in the main paper. The increase in parameters of the U-Net due to this modification is approximately 65K, which is a 0.008\% increase on top of the parameters of the base U-Net (~860M parameters). 

To provide a comprehensive comparison between the 4-channel and 16-channel VAEs, we have analyzed the Image Reconstruction capabilities of both the VAEs. Table ~\ref{tab:vae_comp} shows the Quantitative results, while Figure ~\ref{fig:vae_recon} shows the Qualitative results. It can be observed from Figure ~\ref{fig:vae_recon} that the difference in Generative quality is marginal (almost identical reconstruction). However, our experiments demonstrate that the expanded latent space in U-Net leads to improved reconstruction fidelity for super-resolution tasks (see Figure~\ref{fig:sr_abalation}).

Algorithm ~\ref{alg:latent_expansion} summarizes the overall procedure for constructing the sixteen-channel latent diffusion backbone used as the starting point of our framework. 

\begin{algorithm*}[t]
\caption{Latent Capacity Expansion from 4 to 16 Channels}
\label{alg:latent_expansion}
\KwIn{Dataset $\mathcal{D}$, teacher VAE $(E_t,D_t)$, SD VAE $(E_s,D_s)$}
Modify encoder $E_s$ to produce 16-channel latents\;

\For{each image $x \in \mathcal{D}$}{
$z_t = E_t(x)$\;
$z_s = E_s(x)$\;
Compute latent distillation loss $\|z_s - z_t\|_2^2$\;
Reconstruct $\hat{x} = D_s(z_s)$\;
Compute reconstruction loss $\|x-\hat{x}\|_1$\;
Update $(E_s,D_s)$\;
}

Modify diffusion U-Net to accept 16-channel latents\;
Jointly fine-tune VAE and U-Net with diffusion objective\;

\KwOut{16-channel latent diffusion backbone}
\end{algorithm*}

\section{Compact Diffusion Backbone Discovery}

After constructing the 16-channel diffusion backbone, we design a library of lightweight surrogates for each block as presented in Figure 2 of the main paper. We calibrate the surrogate blocks by distilling the knowledge from the base block using 25\% of the training data. Once the library of surrogates is ready, we perform an architecture search to identify a compact diffusion model that preserves the generative capability of the base network while reducing computational complexity. The architecture search produces multiple candidates that satisfy the $\epsilon$-constraint on $f_{\text{acc}}(\mathbf{a})$ (see Eq. 11 of the main paper), while offering different levels of parameter reduction. As described in Eq. 15 of the main paper, we performed BO on the grid of $\{\epsilon_1 = 0.001, \epsilon_2 = 0.003, \epsilon_3 = 0.005, \epsilon_4 = 0.007\}$.

Table ~\ref{tab: candidates} summarizes the parameter count, different levels of $\epsilon = f_{\text{acc}}(\mathbf{a})$ for which BO was run, and the corresponding FID metric, evaluated on the calibration set by treating the base model's outputs as Ground Truth. These results illustrate the trade-off between model compactness and reconstruction quality.

\begin{table*}[]
\centering
\caption{Candidate architectures obtained during 4 different runs of search using $\epsilon$-constrained Bayesian Optimization based on 4 different thresholds of $\epsilon$ or $f_{\text{acc}}(\mathbf{a})$.} 
\label{tab: candidates}
\begin{tabular}{cccc}
\toprule
Model        & Parameters & $\epsilon = f_{\text{acc}}(\mathbf{a})$ & FID on calib. set \\
\midrule
Candidate 1     & 173M  & 0.001 & 18.3           \\
Candidate 2    & 161M   & 0.003 & 19.4           \\
Candidate 3     & 146M   & 0.005 & 22.9           \\
Candidate 4     & 130M   & 0.007 & 22.6           \\
\bottomrule
\end{tabular}
\end{table*}

\begin{figure}
    \centering
    \includegraphics[width=0.5\linewidth]{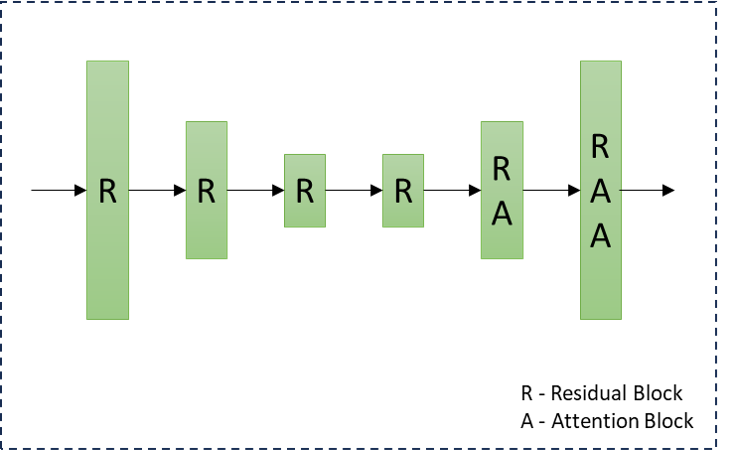}
    \caption{Architecture of the Compact U-Net (Candidate-4 in Table~\ref{tab: candidates}). As opposed to 4 Encoders, 1 Middle block, and 4 Decoders in the 16-channel backbone based on SD-1.5, which can be described as [RARA-RARA-RARA-RR]-[RAR]-[RRR-RARA-RARA-RARA], this compact model has only 3 encoders and 3 decoders with the configuration: [R-R-R]-[R-RA-RAA], where R and A are Residual and Attention blocks.}
    \label{fig:unet}
\end{figure}

\begin{algorithm*}[t]
\caption{Conditional Diffusion Training for Super-Resolution}
\label{alg:algo3}
\textbf{Input:} Compact diffusion backbone $\mathcal{M}_{\phi}$, HR dataset $\mathcal{D}$, VAE encoder $E(\cdot)$, diffusion schedule $\{\alpha_t\}_{t=1}^{T}$  

\textbf{Output:} Conditional SR diffusion model $\epsilon_{\phi}$  

1: \textbf{for} each HR image $x_h \in \mathcal{D}$ \textbf{do}  

2: \quad Generate LR image $x_l$ by bicubic downsampling with scale $3\times$  

3: \quad Encode HR image to latent space  
\[
z_h = E(x_h)
\]

4: \quad Upsample LR image and encode to obtain conditioning latent  
\[
z_l = E(\text{Upsample}(x_l))
\]

5: \quad Sample timestep $t \sim \{1,\dots,T\}$ and noise $\epsilon \sim \mathcal{N}(0,I)$  

6: \quad Generate noisy latent  
\[
z_t = \sqrt{\alpha_t}z_h + \sqrt{1-\alpha_t}\epsilon
\]

7: \quad Predict noise using conditional diffusion backbone  
\[
\hat{\epsilon} = \epsilon_{\phi}(z_t,t,z_l)
\]

8: \quad Update $\phi$ by minimizing  
\[
\mathcal{L}_{SR} = \|\epsilon - \hat{\epsilon}\|_2^2
\]

9: \textbf{end for}

\textbf{return} trained conditional SR diffusion model

\end{algorithm*}

Among the candidate models, to satisfy our budget constraints, we select the architecture with the fewest parameters (130M). Further, to improve the generative capability of the candidate model, it is finetuned end-to-end using the standard diffusion denoising objective as described in Eq. 16 in the main manuscript. Figure~\ref{fig:unet} shows the architecture of the compact U-Net model with 130M parameters. This architecture is then adopted as the compact diffusion backbone UNet for the subsequent super-resolution training stage.

\section{Adapting the Compact Diffusion Backbone for One-Step Super-Resolution}

\label{subsec:training_details}

In addition to the methodology described briefly in the main paper due to space constraints, we provide here the algorithms for the Two-stage training pipeline for finetuning the Generative model to the Super-Resolution model. Algorithm ~\ref{alg:algo3} explains the step-by-step procedure of converting the generative compact backbone into the Super-Resolution expert, and Algorithm ~\ref{algo4} provides the details of step distillation, where we distill the knowledge from the Multi-step teacher into a single DDIM step compact U-Net.

\begin{algorithm*}[t]
\caption{One-Step Distillation for Efficient Super-Resolution}
\label{algo4}
\textbf{Input:} Trained diffusion SR model $\epsilon_{\phi}$, dataset $\mathcal{D}$, VAE encoder $E(\cdot)$, decoder $D(\cdot)$  

\textbf{Output:} One-step super-resolution model $G_{\psi}$  

1: \textbf{for} each LR–HR pair $(x_l,x_h)$ \textbf{do}  

2: \quad Encode HR image to latent representation  
\[
z_h = E(x_h)
\]

3: \quad Encode LR image to conditioning latent  
\[
z_l = E(\text{Upsample}(x_l))
\]

4: \quad Sample initial noisy latent  
\[
z_T \sim \mathcal{N}(0,I)
\]

5: \quad Run teacher diffusion model to obtain denoised latent  
\[
z_0^{T} = \text{Diffusion}(\epsilon_{\phi}, z_T, z_l)
\]

6: \quad Predict one-step latent using student generator  
\[
\hat{z}_0 = G_{\psi}(z_T,z_l)
\]

7: \quad Update $\psi$ by minimizing  
\[
\mathcal{L}_{distill} = \| \hat{z}_0 - z_0^{T} \|_2^2
\]

8: \textbf{end for}

9: Decode predicted latent to produce final SR image  
\[
\hat{x}_h = D(\hat{z}_0)
\]

\textbf{return} trained one-step SR model $G_{\psi}$

\end{algorithm*}

\begin{figure}[ht]
    \centering
    \includegraphics[width=1\linewidth]{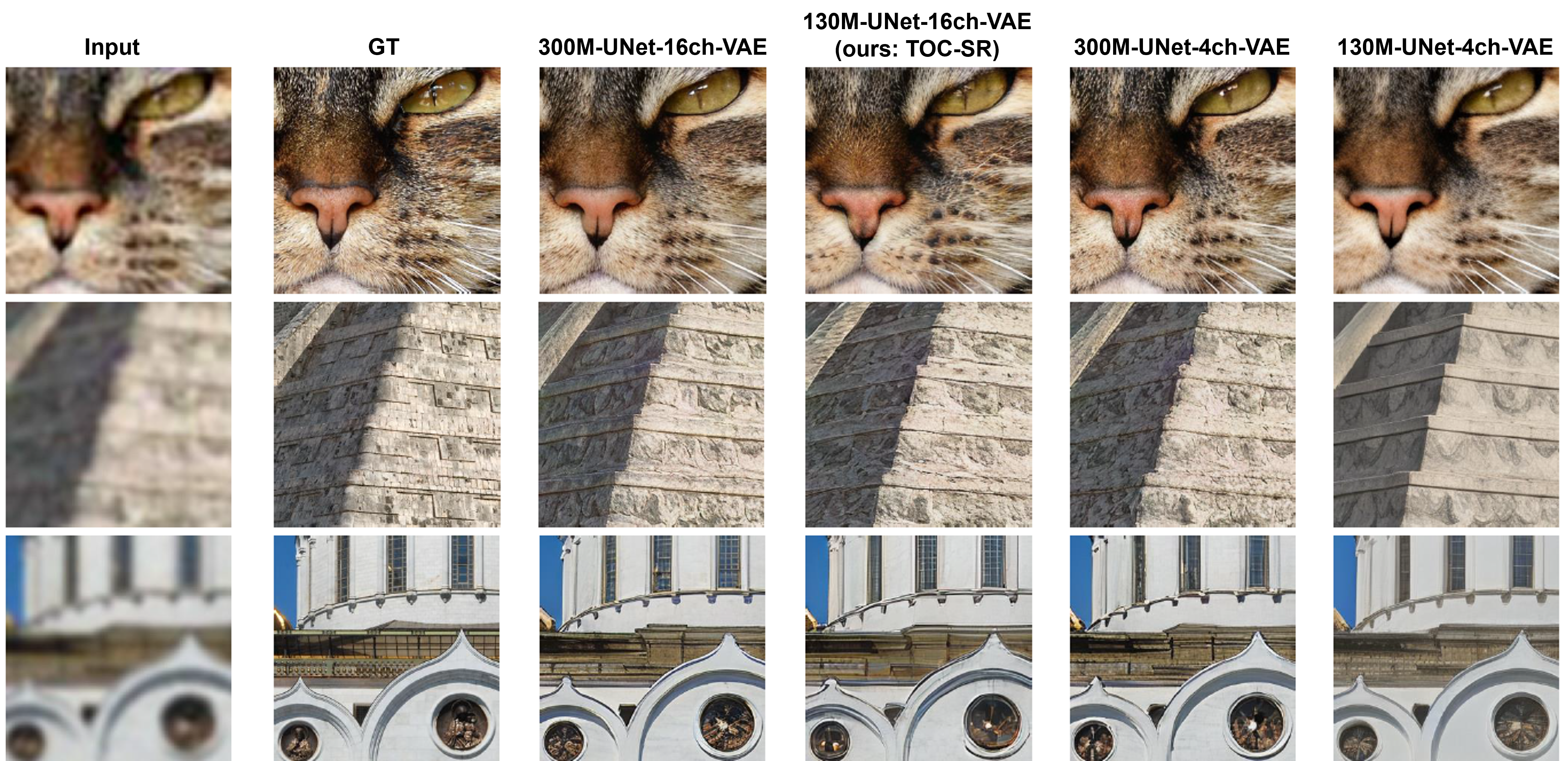}
    \caption{
\textbf{Ablation: impact of latent capacity on SR quality.}
Visual comparison showing TOC-SR against more complex variants trained with the same pipeline.
The 16-channel latent produces sharper, more faithful reconstructions than reduced-latent variants, consistent with the quantitative trends in Table~\ref{table:sr3}.
}
    \label{fig:sr_abalation}
\end{figure}

\begin{table*}[]
\centering
\caption{\textbf{Ablation on VAE latent channels and U-Net capacity (DIV-2K Val, $\times4$).}
We compare 16-channel vs.\ 4-channel VAE latents and 130M vs.\ 300M U-Net capacity under the same training pipeline.
Latent capacity yields the largest quality gains, while increasing U-Net size provides smaller improvements and may not always improve perceptual metrics. Color-coding: red is best, followed by blue and green.}
\label{table:sr3}
\resizebox{\textwidth}{!}{
\begin{tabular}{llllllllll}
\hline
Method                     & PSNR$\uparrow$    & LPIPS$\downarrow$   & FID$\downarrow$  & NIQE$\downarrow$ & MUSIQ$\uparrow$ & Steps & GMACs \\ \hline

300M-UNet-16ch-VAE        & \textcolor{red}{24.73}  & \textcolor{blue}{0.257}  & \textcolor{green}{24.44} & \textcolor{red}{4.32} & \textcolor{red}{71.65} & 1   & 4234        \\ 

 130M-UNet-16ch-VAE  (ours)   &  \textcolor{blue}{24.26 }    &       \textcolor{red}{0.242}         &  \textcolor{red}{22.36}    &  \textcolor{blue}{4.56}  &  \textcolor{blue}{70.14} &    1   &   4145       \\ 

300M-UNet-4ch-VAE     & \textcolor{green}{23.75}  & 0.302  & \textcolor{blue}{23.90} & \textcolor{green}{4.73} & 65.67 & 1    & 1960         \\ 

130M-UNet-4ch-VAE               & 23.24  & \textcolor{green}{0.291}  & 27.37 & 5.1 & \textcolor{green}{66.53} & 1    & 1871         \\ \hline

\end{tabular}
}
\end{table*} 

\section{Ablations on Super-Resolution task}

We ablate two key factors under the same training pipeline: (i) the VAE latent dimensionality (16-channel vs.\ 4-channel), and (ii) the diffusion U-Net capacity (130M vs.\ 300M parameters).
Figure~\ref{fig:sr_abalation} provides qualitative evidence, and Table~\ref{table:sr3} reports quantitative comparisons on DIV2K-Val.

\paragraph {} \textbf{Effect of latent dimensionality (16ch vs.\ 4ch VAE).}
Reducing the latent space from 16 to 4 channels leads to a consistent degradation in both fidelity and perceptual quality in SR task.
For the 130M U-Net, switching from 16ch to 4ch reduces PSNR from 24.26 to 23.24 and worsens LPIPS from 0.242 to 0.291.
A similar drop is observed for the 300M model (24.73$\rightarrow$23.75 PSNR; 0.257$\rightarrow$0.302 LPIPS).
These results indicate that latent capacity is a primary bottleneck for SR quality: a higher-dimensional latent better preserves fine structures and enables the denoiser to reconstruct high-frequency details more faithfully.

\paragraph{}\textbf{Effect of U-Net capacity (130M vs.\ 300M) under fixed latent.}
Increasing the U-Net size improves PSNR (e.g., 24.26$\rightarrow$24.73 for 16ch VAE), but the gains are smaller than those obtained by using a stronger latent representation.
Moreover, the 130M+16ch configuration achieves better perceptual quality than the larger 300M+16ch variant, with lower LPIPS (0.242 vs.\ 0.257) and improved FID (22.36 vs.\ 24.44).
This suggests that TOC-SR benefits more from a task-aligned, higher-capacity latent than from simply scaling the denoiser.

\paragraph{}\textbf{Takeaway.}
Overall, the best quality--efficiency trade-off is achieved by TOC-SR (130M U-Net + 16ch VAE), validating our design choice of combining a compact denoiser with a distilled 16-channel VAE tailored for super-resolution.

\end{document}